# *Samanantar*: The Largest Publicly Available Parallel Corpora Collection for 11 Indic Languages

**Gowtham Ramesh**[1*] **Sumanth Doddapaneni**[1*] **Aravinth Bheemaraj**[2,5]
**Mayank Jobanputra**[3] **Raghavan AK**[4] **Ajitesh Sharma**[2,5] **Sujit Sahoo**[2,5]
**Harshita Diddee**[4] **Mahalakshmi J**[4] **Divyanshu Kakwani**[3,4] **Navneet Kumar**[2,5]
**Aswin Pradeep**[2,5] **Srihari Nagaraj**[2,5] **Kumar Deepak**[2,5] **Vivek Raghavan**[5]
**Anoop Kunchukuttan**[4,6] **Pratyush Kumar**[1,3,4] **Mitesh Shantadevi**[†] **Khapra**[1,3,4‡]
[1]RBCDSAI, [2]Tarento Technologies, [3]IIT Madras
[4]AI4Bharat, [5]EkStep Foundation, [6]Microsoft

## Abstract

We present *Samanantar*, the largest publicly available parallel corpora collection for Indic languages. The collection contains a total of 49.7 million sentence pairs between English and 11 Indic languages (from two language families). Specifically, we compile 12.4 million sentence pairs from existing, publicly-available parallel corpora, and additionally mine 37.4 million sentence pairs from the web, resulting in a 4× increase. We mine the parallel sentences from the web by combining many corpora, tools, and methods: (a) web-crawled monolingual corpora, (b) document OCR for extracting sentences from scanned documents, (c) multilingual representation models for aligning sentences, and (d) approximate nearest neighbor search for searching in a large collection of sentences. Human evaluation of samples from the newly mined corpora validate the high quality of the parallel sentences across 11 languages. Further, we extract 83.4 million sentence pairs between all 55 Indic language pairs from the English-centric parallel corpus using English as the pivot language. We trained multilingual NMT models spanning all these languages on *Samanantar* which outperform existing models and baselines on publicly available benchmarks, such as FLORES, establishing the utility of *Samanantar*. Our data and models are available publicly at Samanantar and we hope they will help advance research in NMT and multilingual NLP for Indic languages.

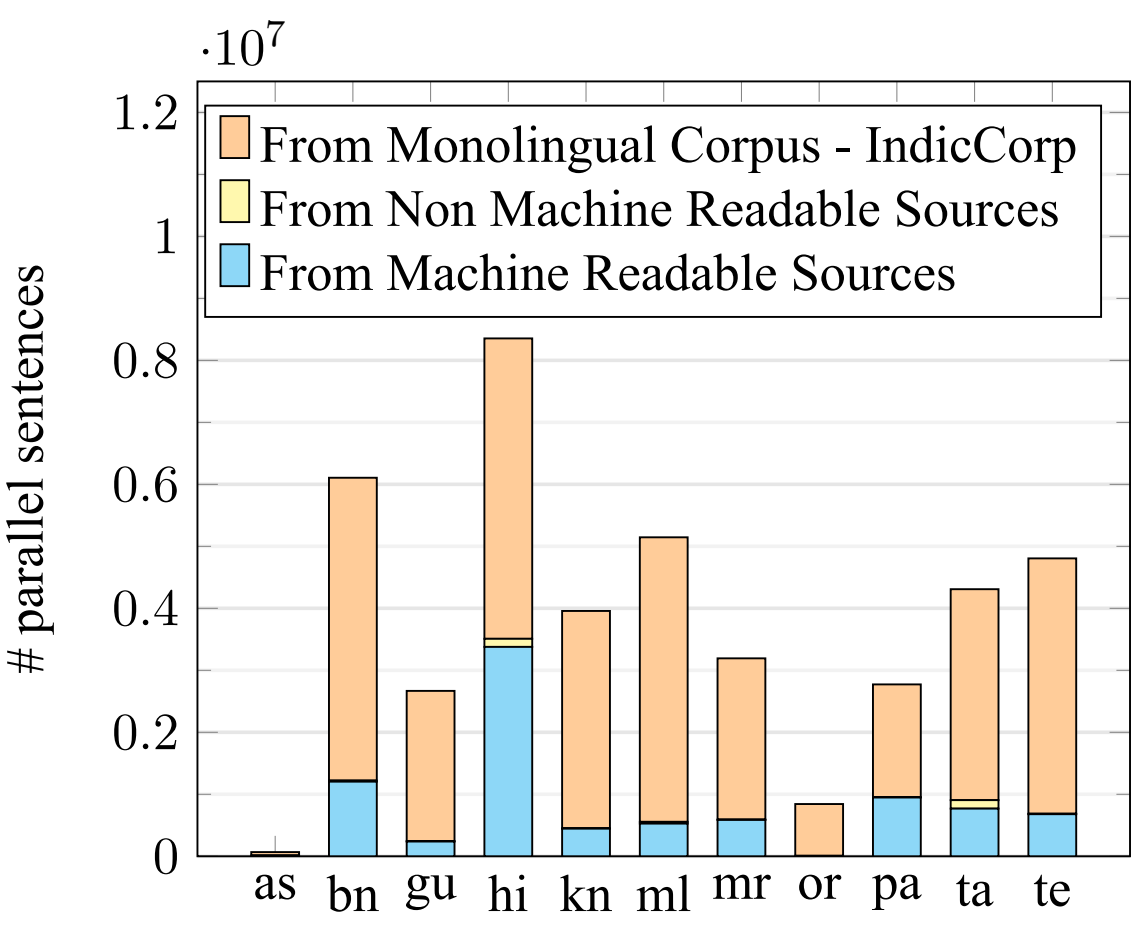

Figure 1: Total number of newly mined En-X parallel sentences in *Samanantar* from different sources.

## 1 Introduction

The advent of deep-learning (DL) based neural encoder-decoder models has lead to significant progress in machine translation (MT) (Bahdanau et al., 2015; Wu et al., 2016; Sennrich et al., 2016b,a; Vaswani et al., 2017). While this has been favorable for resource-rich languages, there has been limited benefit for resource-poor languages which lack parallel corpora, monolingual corpora and evaluation benchmarks (Koehn and Knowles, 2017). Multilingual models can improve performance on resource-poor languages via transfer learning from resource-rich languages (Firat et al., 2016; Johnson et al., 2017b; Kocmi and Bojar, 2018), more so when the resource-rich and resource-poor languages are related (Nguyen and Chiang, 2017; Dabre et al., 2017). However, it is difficult to achieve this with limited in-language data (Guzmán et al., 2019), particularly when an entire group of related languages is low-resource making transfer-learning infeasible.

A case in point is that of languages from the Indian subcontinent, a very linguistically diverse region. India has 22 constitutionally listed languages spanning 4 major language families. Other countries in the subcontinent also have their share of widely spoken languages. These languages are closely related both genetically and through contact, which led to significant sharing of vocabulary and linguistic features (Emeneau, 1956). These languages account for a collective speaker base of over 1 billion speakers. The demand for quality, publicly available translation systems in a multilingual society like India is obvious. However, there is very limited publicly available parallel data for Indic languages. Given this situation, an obvious question to ask is: *What does it take to improve MT on the large set of related low-resource Indic languages?* The answer is straightforward: *create large parallel datasets and train proven DL models*. However, collecting new data with manual translations at the scale necessary to train large DL models would be slow and expensive. Instead, several recent works have proposed mining parallel sentences from the web (Schwenk et al., 2019a, 2020; El-Kishky et al., 2020). The representation of Indic languages in these works is however poor (e.g., CCMatrix contains parallel data for only 6 Indic languages). In this work, we aim to significantly increase the amount of parallel data on Indic languages by combining the benefits of many recent contributions: large Indic monolingual corpora (Kakwani et al., 2020; Ortiz Suarez et al., 2019), accurate multilingual representation learning (Feng et al., 2020; Artetxe and Schwenk, 2019), scalable approximate nearest neighbor search (Johnson et al., 2017a; Subramanya et al., 2019; Guo et al., 2020), and optical character recognition (OCR) of Indic scripts in rich text documents. By combining these methods, we propose different pipelines to collect parallel data from three different types of sources: (a) non machine readable sources like scanned parallel documents, (b) machine-readable sources like news websites with multilingual content, (c) IndicCorp (Kakwani et al., 2020), the largest corpus of monolingual data for Indic languages.

* The first two authors have contributed equally.
† Dedicated to the loving memory of my grandmother.
‡ Corresponding author: miteshk@cse.iitm.ac.in

| Source | en-as | en-bn | en-gu | en-hi | en-kn | en-ml | en-mr | en-or | en-pa | en-ta | en-te | Total |
|---|---|---|---|---|---|---|---|---|---|---|---|---|
| Existing Sources | 108 | 3,496 | 611 | 2,818 | 472 | 1,237 | 758 | 229 | 631 | 1,456 | 593 | 12,408 |
| New Sources | 34 | 5,109 | 2,457 | 7,308 | 3,622 | 4,687 | 2,869 | 769 | 2,349 | 3,809 | 4,353 | 37,366 |
| Total | 141 | 8,605 | 3,068 | 10,126 | 4,094 | 5,924 | 3,627 | 998 | 2,980 | 5,265 | 4,946 | 49,774 |
| *Increase Factor* | 1.3 | 2.5 | 5 | 3.6 | 8.7 | 4.8 | 4.8 | 4.4 | 4.7 | 3.6 | 8.3 | 4 |

Table 1: Summary statistics of the Samanatar corpus. All numbers are in thousands.

Combining existing datasets and the new datasets that we mine from the above mentioned sources, we present *Samanantar*[1] - the largest publicly available parallel corpora collection for Indic languages. *Samanantar* contains $\sim$ 49.7M parallel sentences between English and 11 Indic languages, ranging from 141K pairs between English-Assamese to 10.1M pairs between English-Hindi. Of these 37.4M pairs are newly mined as a part of this work whereas 12.4M are compiled from existing sources. Thus, the newly mined data is about 3 times the existing data. Table 1 shows the language-wise statistics. Figure 1 shows the relative contribution of different sources from which new parallel sentences were mined. The largest contributor is data mined from IndicCorp which accounts for 67% of the total corpus. *From this English-centric corpus, we mine 83.4M parallel sentences between the 55 ($\binom{11}{2}$) Indic language pairs using English as the pivot.* To evaluate the quality of the mined sentences we collect human judgments from 38 annotators for a total of 9,566 sentence pairs across 11 languages. The annotations attest to the high quality of the mined parallel corpus and validate our design choices.

To evaluate if *Samanantar* advances the state of the art for Indic NMT, we train a multilingual model, called *IndicTrans*, using *Samanantar*. We compare *IndicTrans*, with (a) commercial translation systems (Google, Microsoft), (b) publicly available translation systems OPUS-MT (Tiedemann and Thottingal, 2020a), mBART50 (Tang et al., 2020), CVIT-ILMulti (Philip et al., 2020), and (c) models trained on all existing sources of parallel data between Indic languages. Across multiple publicly available test sets spanning 10 Indic languages, we observe that *IndicTrans* performs better than all existing open source models and even outperforms commercial systems on many benchmarks, thereby establishing the utility of *Samanantar*.

The three main contributions of this work, *viz.*, (i) *Samanantar*, the largest collection of parallel corpora for Indic languages, (ii) *IndicTrans*, a multilingual model for translating from En-Indic and Indic-En, and (iii) human judgments on cross-lingual textual similarity for about 9,566 sentence pairs will be made publicly available.

[1] *Samanantar* in Sanskrit means semantically similar

## 2 *Samanantar*: A Parallel Corpus for Indic Languages

*Samanantar* contains parallel sentences between English and 11 Indic languages, *viz*., Assamese (as), Bengali (bn), Gujarati (gu), Hindi (hi), Kannada (kn), Malayalam (ml), Marathi (mr), Odia (or), Punjabi (pa), Tamil (ta) and Telugu (te). In addition, it also contains parallel sentences between the $\binom{11}{2} = 55$ Indic language pairs obtained by pivoting through English (en). To build this corpus, we first collated all existing public sources of parallel data for Indic languages that have been released over the years, as described in Section 2.1. We then expand this corpus further by mining parallel sentences from three types of sources from the web as described in sections 2.2 to 2.4.

### 2.1 Collation from existing sources

We first briefly describe the existing sources of parallel sentences for Indic languages. The *Indic NLP Catalog*[2] helped identify many of these sources. Recently, the WAT 2021 MultiIndicMT shared task (Nakazawa et al., 2021) also compiled many existing Indic language parallel corpora.

Some sentence aligned corpora were collected from OPUS[3] (Tiedemann, 2012) on **21 March 2021**. These include localization data (GNOME, KDE4, Ubuntu, Mozilla-I10n), religious text (JW300 (Agić and Vulić, 2019), Bible-eudin (Christodouloupoulos and Steedman, 2015), Tanzil), GlobalVoices, OpenSubtitles (Lison and Tiedemann, 2016), TED2020 (Reimers and Gurevych, 2020), WikiMatrix (Schwenk et al., 2019a), Tatoeba and ELRC_2922.

We also collated parallel data from the following non-OPUS sources (URLs can be found in the *Indic NLP Catalog*): ALT (Riza et al., 2016), BanglaNMT (Hasan et al., 2020), CVIT-PIB (Philip et al., 2020), IITB (Kunchukuttan et al., 2018), MTEnglish2Odia, NLPC, OdiEnCorp 2.0 (Parida et al., 2020), PMIndia V1 (Haddow and Kirefu, 2020), SIPC (Post et al., 2012), TICO19 (Anastasopoulos et al., 2020), UFAL (Ramasamy et al., 2012), URST (Shah and Bakrola, 2019) and WMT (Barrault et al., 2019) provided training set for en-gu.

As shown in Table 1, these sources[4] collated together result in a total of 12.4M parallel sentences (after removing duplicates) between English and 11 Indic languages. It is interesting that no publicly available MT system has been trained using parallel data from all these existing sources.

We observed that some existing sources, such as JW300, were extremely noisy containing many sentence pairs which were not translations of each other. However, we chose not to clean/post-process any of the existing sources, beyond what was already done by the public repositories that released these datasets. As future work, we plan to study different data filtering (Junczys-Dowmunt, 2018) and data sampling techniques (Bengio et al., 2009) and their impact on the performance of the NMT model being trained. For example, we could sort the sources by their quality and feed sentences from only very high quality sources during the later epochs while training the model.

### 2.2 Mining parallel sentences from machine readable comparable corpora

We identified several news websites which publish articles in multiple Indic languages (see Table 2). For a given website, the articles across languages are not necessarily translations of each other. However, content within a given date range is often similar as the sources are India-centric with a focus on local events, personalities, advisories, *etc*. For example, news about guidelines for CoViD-19 vaccination get published in multiple Indic languages. Even if such a news article in Hindi is not a sentence-by-sentence translation, it may contain some sentences which are accidentally or intentionally parallel to sentences from a corresponding English article. Hence, we consider such news websites to be good sources of parallel sentences.

We also identified some sources from education domain - NPTEL[5], Coursera[6], Khan Academy[7] and some science Youtube channels which provide educational videos with parallel human translated subtitles in different Indic languages.

We use the following steps to extract parallel sentences from the above sources:

**Article Extraction.** For every news website,

[2] https://github.com/AI4Bharat/indicnlp_catalog

[3] URLs to the original sources can be found on the OPUS website: https://opus.nlpl.eu

[4] We have not included CCMatrix(Schwenk et al., 2020) and CCAligned(El-Kishky et al., 2020) in the current version of Samanantar. CCMatrix is not publicly available and CCAligned has been criticised by some recent work (Caswell et al., 2021).

[5] https://nptel.ac.in

[6] https://www.coursera.org/

[7] https://www.khanacademy.org

| Mykhel | DD national + sports | Punjab govt | Pranabmukherjee | Catchnews | Nptel |
|---|---|---|---|---|---|
| Drivespark | Financial Express | Gujarati govt | General_corpus | Kolkata24x7 | Wikipedia |
| Good returns | Zeebiz | Business Standard | NewsOnAir | Asianetnews | Coursera |
| Indian Express | Sakshi | The Wire | Nouns_dictionary | YouTube science channels | |
| The times of india | Marketfeed | The Bridge | PIB | Prothomalo | |
| Nativeplanet | Jagran | The Better India | PIB_archives | Khan_academy | |

Table 2: Machine Readable Sources in *Samanantar*

we build custom extractors using BeautifulSoup[8] or Selenium[9] to extract the main article content. For NPTEL, Youtube science channels and Khan Academy, we use youtube-dl[10] to collect Indic and English subtitles for every video. We skip the auto-generated youtube captions to ensure that we only get high quality translations. We collected subtitles for all available courses/videos on **March 7th, 2021**. For Coursera, we identify courses which have manually created Indic and English subtitles and then use coursera-dl[11] to extract these subtitles.

**Tokenisation.** We split the main content of the articles into sentences using the Indic NLP Library[12] (Kunchukuttan, 2020), with a few additional heuristics to account for Indic punctuation characters, sentence delimiters and non-breaking prefixes.

**Parallel Sentence Extraction.** At the end of the above step, we have sentence tokenised articles in English and a target language (say, Hindi). Further, all these news websites contain metadata based on which we can cluster the articles according to the month in which they were published (say, January 2021). We assume that to find a match for a given Hindi sentence we only need to consider all English sentences which belong to articles published in the same month as the article containing the Hindi sentence. This is a reasonable assumption as content of news articles is temporal in nature. Note that such clustering based on dates is not required for the education sources as there we can find matching sentences in bilingual captions belonging to the same video.

Let $S = \{s_1, s_2, \dots, s_m\}$ be the set of all sentences across all English articles in a particular month (or in the English caption file corresponding to a given video). Similarly, let $T = \{t_1, t_2, \dots, t_n\}$ be the set of all sentences across all Hindi articles in that same month (or in the Hindi caption file corresponding to the same video). Let $f(s, t)$ be a scoring function which assigns a score indicating how likely it is that $s \in S, t \in T$ form a translation pair. For a given Hindi sentence $t_i \in T$, the matching English sentence can be found as:

$$s^* = \arg\max_{s \in S} f(s, t_i)$$

We chose $f$ to be the cosine similarity function on embeddings of $s$ and $t$. We compute these embeddings using LaBSE (Feng et al., 2020) which is a state-of-the-art multilingual sentence embedding model that encodes text from different languages into a shared embedding space. We refer to the cosine similarity between the LaBSE embeddings of $s, t$ as the LaBSE Alignment Score (LAS).

**Post Processing.** Using the above described process, we find the top matching English sentence, $s^*$, for every Hindi sentence, $t_i$. We now apply a threshold and select only those pairs for which the cosine similarity is greater than a threshold $t$. Across different sources we found 0.75 to be a good threshold. We refer to this as the *LAS threshold*. Next, we remove duplicates in the data. We consider two pairs $(s_i, t_i)$ and $(s_j, t_j)$ to be duplicate if $s_i = s_j$ *and* $t_i = t_j$. We also remove any sentence pair where the English sentence is less than 4 words. Lastly, we use a language identifier[13] and eliminate pairs where the language identified for $s_i$ or $t_i$ does not match the intended language.

### 2.3 Mining parallel sentences from non-machine readable comparable corpora

While web sources are machine readable, there are official documents that are generated which are not always machine readable. For example, proceedings of the legislative assemblies of different Indian states in English as well as the official language of the state are published as PDFs. In this work, we considered 3 such public sources: (a) documents from Tamil Nadu government[14] (en-ta), (b) speeches from Bangladesh Parliament[15] and West Bengal Legislative Assembly[16] (en-bn), and (c) speeches from Andhra Pradesh[17] and Telangana Legislative Assemblies[18] (en-te). Most of these documents either contained scanned images of the original document or contained proprietary encodings (non-UTF8) due to legacy issues. As a result, standard PDF parsers cannot be used to extract text from them. We use the following pipeline for extracting parallel sentences from such sources.

[8] https://www.crummy.com/software/BeautifulSoup
[9] https://pypi.org/project/selenium
[10] https://github.com/tpikonen/youtube-dl
[11] https://github.com/coursera-dl/coursera-dl
[12] https://github.com/anoopkunchukuttan/indic_nlp_library
[13] https://github.com/aboSamoor/polyglot

**Optical Character Recognition (OCR).** We used Google's Vision API, which supports English as well as the 11 Indic languages considered, to extract text from each document.

**Tokenisation.** We use the same tokenisation process as described in the previous section on the extracted text with extra heuristics to merge an incomplete sentence at the bottom of one page with an incomplete sentence at the top of the next page.

**Parallel Sentence Extraction.** Unlike the previous section, we have exact information about which documents are parallel. This information is typically encoded in the URL of the document itself (e.g., https://tn.gov.in/**en**/budget.pdf and https://tn.gov.in/**ta**/budget.pdf). Hence, for a given Tamil sentence, $t_i$ we only need to consider the sentences $S = \{s_1, s_2, \ldots, s_m\}$ which appear in the corresponding English article. For a given $t_i$, we identify the matching sentence, $s^*$, from the candidate set $S$, using LAS as described in sec. 2.2.

**Post-Processing.** We use the same post-processing as described in section 2.2.

### 2.4 Mining parallel sentences from web scale monolingual corpora

Recent works (Schwenk et al., 2019b; Feng et al., 2020) have shown that it is possible to align parallel sentences in large monolingual corpora (e.g., CommonCrawl) by computing the similarity between them in a shared multilingual embedding space. In this work, we consider *IndicCorp* (Kakwani et al., 2020), the largest collection of monolingual corpora for Indic languages (ranging from 1.39M sentences for Assamese to 100.6M sentences for English). The idea is to take an Indic sentence and find its matching En sentence from a large collection of En sentences. To perform this search efficiently, we use FAISS (Johnson et al., 2017a) which does efficient indexing, clustering, semantic matching, and retrieval of dense vectors as explained below.

**Indexing.** We compute the sentence embedding using LaBSE for all English sentences in *IndicCorp*. We create a FAISS index where these embeddings are stored in $100k$ clusters. We use Product Quantization (Jégou et al., 2011) to reduce the space required to store these embeddings by quantizing the 786 dimensional LaBSE embedding into a $m$ dimensional vector ($m = 64$) where each dimension is represented using an 8-bit integer value.

**Retrieval.** For every Indic sentence (say, Hindi sentence) we first compute the LaBSE embedding and then query the FAISS index for its nearest neighbor based on normalized inner product (i.e., cosine similarity). FAISS first finds the top-$p$ clusters by computing the distance between each of the cluster centroids and the given Hindi sentence. We set the value of $p$ to 1024. Within each of these clusters, FAISS searches for the nearest neighbors. This retrieval is highly optimized to scale. In our implementation, on average we were able to perform 1100 nearest neighbourhood searches per second on the index containing 100.6M En sentences.

**Recomputing cosine similarity.** Note that FAISS computes cosine similarity on the quantized vectors (of dimension $m = 64$). We found that while the relative ranking produced by FAISS is good, the similarity scores on the quantized vectors vary widely and do not accurately capture the cosine similarity between the original 768d LaBSE embeddings. Hence, it is difficult to choose an appropriate threshold on the similarity of the quantized vector. However, the relative ranking provided by FAISS is still good. For example, for all the 100 query Hindi sentences that we analysed, FAISS retrieved the correct matching English sentence from an index of 100.6 M sentences at the top-1 position. Based on this observation, we follow a two-step approach: First, we retrieve the top-1 matching sentence from FAISS using the quantized vector. Then, we compute the LAS between the full LaBSE embeddings of the retrieved sentence pair. On the computed LAS, we apply a LAS threshold of 0.80 (slightly higher than that the one used for comparable sources described earlier) for filtering. This modified FAISS mining, combining quantized vectors for efficient searching and full embeddings from LaBSE for accurate thresholding, was crucial for mining a large number of parallel sentences.

[14] https://www.tn.gov.in/documents/deptname
[15] http://www.parliament.gov.bd
[16] http://www.wbassembly.gov.in
[17] https://www.aplegislature.org, https://www.apfinance.gov.in
[18] https://finance.telangana.gov.in

**Post-processing.** We follow the same post-processing steps as described in Section 2.2.

We also used the above process to extract parallel sentences from Wikipedia by treating it as a collection of monolingual sentences in different languages. We were able to mine more parallel sentences using this approach as opposed to using Wikipedia's interlanguage links for article alignment followed by inter-article parallel sentence mining.

Note that we chose this LaBSE based alignment method over existing methods like Vecalign (Thompson and Koehn, 2019) and Bleualign (Sennrich and Volk, 2011) as these methods assume/require parallel documents. However, for IndicCorp, such a parallel alignment of documents is not available and may not even exist. Further, LaBSE is trained on 17 billion monolingual sentences and 6 billion bilingual sentence pairs using from 109 languages including all the 11 Indic languages considered in this work. The authors have shown that it produces state of the art results on multiple parallel text retrieval tasks and is effective even for low-resource languages. Given these advantages of LaBSE embeddings and to have a uniform scoring mechanism (i.e., LAS) across sources, we use the same LaBSE based mechanism for mining parallel sentences from all the sources that we considered.

### 2.5 Mining Inter-Indic Language Corpora

So far, we have discussed mining parallel corpora between English and Indic languages. Following Freitag and Firat (2020) and Rios et al. (2020), we now use English as a pivot to mine parallel sentences between Indic languages from all the English-centric corpora described earlier in this section. Most of the sources that we crawled data from for creating *Samanantar* were English-centric, *i.e.*, they contain data in English and one or more Indian languages. Hence we chose English as the pivot language. For example, let $(s^{en}, t^{hi})$ and $(\hat{s}^{en}, t^{ta})$ be mined parallel sentences between en-hi and en-ta respective. If $s^{en} = \hat{s}^{en}$ then we extract $(t^{hi}, t^{ta})$ as a Hindi-Tamil parallel sentence pair. Further, we use a very strict de-duplication criterion to avoid the creation of very similar parallel sentences. For example, if an *en* sentence is aligned to $m$ *hi* sentences and $n$ *ta* sentences, then we would get $mn$ *hi-ta* pairs. We retain only 1 randomly chosen pair out of these $mn$ pairs, since these $mn$ pairs are likely to be similar. We mined 83.4M parallel sentences between the $\binom{11}{2}$ Indic language pairs resulting in a 5.33× increase in publicly available sentence pairs between these languages (see Table 3).

## 3 Analysis of the Quality of the Mined Parallel Corpus

We now describe the intrinsic evaluation of the data that we mined as a part of this work using using the methods described in sections 2.2, 2.3 and 2.4). This evaluation was performed by asking human annotators to estimate cross-lingual Semantic Textual Similarity (STS) of the mined parallel sentences

### 3.1 Annotation Task and Setup

We sampled 9,566 sentence pairs (English and Indic) from the mined data across 11 Indic languages and several sources. The sampling was stratified to have equal number of sentences from three sets:

- *Definite accept*: sentence pairs with LAS larger than 0.1 of the chosen threshold.
- *Marginal accept*: sentence pairs with LAS larger than but within 0.1 of the chosen threshold.
- *Reject*: sentence pairs with LAS smaller than but within 0.1 of the chosen threshold.

The sampled sentences were shuffled randomly such that no ordering is preserved across sources or LAS. We then divided the language-wise sentence pairs into annotation batches of 30 parallel sentences each.

For defining the annotation scores, we refer to the SemEval-2016 Task 1 (Agirre et al., 2016), wherein crosslingual semantic textual similarity is characterised by six ordinal levels ranging from complete semantic equivalence (5) to complete semantic dissimilarity (0). These guidelines were explained to 38 annotators across 11 Indic languages, with a minimum of 2 annotators per language. Each annotator is a native speaker in the language assigned and is also fluent in English. The annotators have experience of 1 to 20 years in working on language tasks, with a mean of 5 years. The annotation task was performed on Google forms: Each form consisted of 30 sentence pairs from an annotation batch. Annotators were shown one sentence pair at a time and were asked to score it in the range of 0 to 5. The SemEval-2016 guidelines were visible to annotators at all times. After annotating 30 parallel sentences, the annotators submitted the form and resumed again with a new form. Annotators were compensated at the rate of Rs 100 to Rs 150 (1.38 to 2.06 USD) per 100 words read.

| | as | bn | gu | hi | kn | ml | mr | or | pa | ta | te | Total |
|---|---|---|---|---|---|---|---|---|---|---|---|---|
| **as** | - | 356 | 142 | 162 | 193 | 227 | 162 | 70 | 108 | 214 | 206 | 1839 |
| **bn** | | - | 1576 | 2627 | 2137 | 2876 | 1847 | 592 | 1126 | 2432 | 2350 | 17920 |
| **gu** | | | - | 2465 | 2053 | 2349 | 1757 | 529 | 1135 | 2054 | 2302 | 16361 |
| **hi** | | | | - | 2148 | 2747 | 2086 | 659 | 1637 | 2501 | 2434 | 19466 |
| **kn** | | | | | - | 2869 | 1819 | 533 | 1123 | 2498 | 2796 | 18168 |
| **ml** | | | | | | - | 1827 | 558 | 1122 | 2584 | 2671 | 19829 |
| **mr** | | | | | | | - | 581 | 1076 | 2113 | 2225 | 15493 |
| **or** | | | | | | | | - | 507 | 1076 | 1114 | 6218 |
| **pa** | | | | | | | | | - | 1747 | 1756 | 11336 |
| **ta** | | | | | | | | | | - | 2599 | 19816 |
| **te** | | | | | | | | | | | - | 20453 |

Table 3: The number of parallel sentences (in thousands) between Indic language pairs. The 'Total' column indicates the aggregate parallel corpus for the language in a row with other Indic languages.

### 3.2 Annotation Results and Discussion

The results of the annotation of the 9,566 sentence pairs and almost 30,000 annotations are shown language-wise in Table 4. Over 85% of the sentence pairs are such that annotators agree within a semantic similarity score of 1 of each other. We make the following key observations from the data.

**Sentence pairs included in *Samanantar* have high semantic similarity.** Overall, the 'All accept' sentence pairs received a mean STS score of 4.27 and a median of 5. On a scale of 0 to 5, where 5 represents perfect semantic similarity, these statistics indicate that annotators rated sentence pairs that are included in *Samanantar* to be of high quality. Furthermore, the chosen LAS thresholds sensitively regulate quality: the 'Definite accept' sentence pairs have a high average STS score of 4.63, which reduces to 3.89 with 'Marginal accept', and significantly falls to 2.94 with the 'Reject' sets.

**LaBSE alignment and annotator scores are moderately correlated.** The Spearman correlation coefficient between LAS and STS is a moderately positive value of 0.37, *i.e.*, sentence pairs with a higher LAS are more likely to be rated to be semantically similar. However, the correlation coefficient is also not very high (say > 0.5) indicating potential for further improvement in learning multilingual representations with LaBSE-like models. Further, the two languages which have the smallest correlation (As and Or) also have the smallest resource sizes, indicating potential for improvement in alignment methods for low-resource languages.

**LaBSE alignment is negatively correlated with sentence length, while annotator scores are not.**

To be consistent across languages, sentence length is computed for the English sentence in each pair. We find that sentence length is negatively correlated with LAS with a Spearman correlation coefficient of -0.35, while it is almost uncorrelated with STS with a Spearman correlation coefficient of -0.04. In other words, pairs with longer sentences are less likely to have high alignment on LaBSE representations.

**Error analysis of mined corpora** For error analysis we considered those sentence pairs as *accurate sentences* which had (a) LAS greater than the threshold, i.e., both marginally accept and definitely accept, and (b) human annotation score greater than or equal to 4. We found that extraction accuracy is 79.5% overall, while the extraction accuracy for Definitely accept bucket is 90.1%. This shows that LAS score based mining and filtering can yield high-quality parallel corpora with high accuracy. In Table 5 we call out different styles of errors for each of the 3 buckets. In Marginally Reject (MR) bucket, we find cases where English and aligned language sentences are different in mean-ing and cannot be treated as parallel sentences altogether. In Marginally Accept (MA) and Definitely Accept (DA) buckets, we find more minor errors, for instance differences in quantity / number and mistaken alignment of special words like Quarter finals (in English) being aligned to Semi finals (in Indic languages).

| Language | Annotation data | | Semantic Textual Similarity score | | | | Spearman correlation coefficient | | |
|---|---|---|---|---|---|---|---|---|---|
| | # Bitext pairs | # Annotations | All accept | Definite accept | Marginal accept | Reject | LAS, STS | LAS, Sentence len | STS, Sentence len |
| Assamese | 689 | 1,972 | 3.52 | 3.86 | 3.11 | 2.18 | 0.25 | -0.39 | 0.19 |
| Bengali | 957 | 3,797 | 4.59 | 4.86 | 4.31 | 3.53 | 0.45 | -0.43 | -0.16 |
| Gujarati | 779 | 2,298 | 4.08 | 4.54 | 3.59 | 2.67 | 0.49 | -0.31 | -0.08 |
| Hindi | 1,276 | 4,616 | 4.50 | 4.84 | 4.14 | 3.15 | 0.48 | -0.18 | -0.12 |
| Kannada | 957 | 2,838 | 4.20 | 4.61 | 3.78 | 2.81 | 0.39 | -0.38 | -0.09 |
| Malayalam | 948 | 2,760 | 4.00 | 4.46 | 3.55 | 2.45 | 0.40 | -0.33 | 0.03 |
| Marathi | 779 | 1,984 | 4.07 | 4.52 | 3.54 | 2.67 | 0.40 | -0.36 | -0.04 |
| Odia | 500 | 1,264 | 4.49 | 4.63 | 4.34 | 4.33 | 0.15 | -0.42 | -0.05 |
| Punjabi | 688 | 2,222 | 4.23 | 4.67 | 3.74 | 2.32 | 0.43 | -0.25 | 0.06 |
| Tamil | 1,044 | 2,882 | 4.29 | 4.62 | 3.95 | 2.57 | 0.35 | -0.40 | -0.14 |
| Telugu | 949 | 2,516 | 4.62 | 4.87 | 4.34 | 3.62 | 0.36 | -0.40 | -0.09 |
| Overall | 9,566 | 29,149 | 4.27 | 4.63 | 3.89 | 2.94 | 0.37 | -0.35 | -0.04 |

Table 4: Results of the annotation task to evaluate the semantic similarity between sentence pairs across 11 languages. Human judgments confirm that the mined sentences (All accept) have a high semantic similarity and with a moderately high correlation between the human judgments and LAS.

| Indian Sentence | English Sentence | LAS | Bucket | Error |
|---|---|---|---|---|
| எனவே, இந்த கொள்கலன்கள் என்ன? | So, what are their strengths? | 0.70 | MR | Should be "So, what are these containers?" |
| ஜொகூர் மாநிலத்தில் வட மேற்கே<br>அமைந்து இருக்கும் இந்த நகரத்தின்<br>மாவட்டமும் மூவார் என்றே அழைக்கப் படுகிறது. | Johor, also spelled as<br>Johore, is a state of Malaysia in<br>the south of the Malay Peninsula. | 0.70 | MR | Should be "Located in the north-western<br>part of the state of Johor, the district of<br>this city is also known as Muwar." |
| ఈ హైదరాబాద్-దుబాయ్ టూర్ హైదరాబాద్‌లోని<br>శంషాబాద్ విమానాశ్రయం నుంచి ప్రారంభమవుతుంది | The flights between Hyderabad and<br>Gorakhpur will begin from 30 April | 0.68 | MR | Should be"This Hyderabad-Dubai<br>tour will start from Shamshabad<br>airport in Hyderabad" |
| இந்த மாவட்டத்தை ஆறு மண்டலங்களாகப்<br>பிரித்துள்ளனர். | the province is divided into ten districts. | 0.81 | MA | Should be "six districts" |
| ఈ నేపథ్యంలో సెన్సెక్స్ సెన్సెక్స్ 511 పాయింట్లకు ఎగియగా, నిఫ్టీ<br>కూడా మద్దతు స్థాయికి ఎగువన స్థిరంగా కొనసాగింది. | The Sensex was trading with gains of 150<br>points, while the Nifty rose 52 points in trade. | 0.76 | MA | Mistake with numbers |
| పారుపల్లి కశ్యప్ కొరియా ఓపెన్ క్వార్టర్<br>ఫైనల్స్‌కు దూసుకెళ్లాడు. | Parupalli Kashyap advcanced to Korea<br>Open semifinals. | 0.88 | DA | semifinals became quarter finals |
| கண்ணின் உட்பகுதியில் யுவெய்டிஸ் எனப்படும் அழற்சி<br>ஏற்படுதல், கண் வலியை ஏற்படுத்தும், குறிப்பாக அதிக<br>ஒளிக்கு ஆளாகும் போது (ஃபோட்டோபோபியா). | Inflammation of the interior portion of the eye,<br>known as uveitis, can cause blurred vision and<br>eye pain, especially when exposed to light<br>(photophobia). | 0.89 | DA | It should be "when exposed to<br>high amounts of light" |

Table 5: Table shows the various errors for different classes in LaBSE based alignment

In summary, the annotation task established that the parallel sentences in *Samanantar* are of high quality and validated the chosen thresholds. The task also established that LaBSE-based alignment should be further improved for low-resource languages (like as, or) and for longer sentences. We will release this parallel dataset and human judgments on the over 9,566 sentence pairs as a dataset for evaluating cross-lingual semantic similarity between English and Indic languages.

## 4 *IndicTrans*: Multilingual, single Indic script models

The languages in the Indian subcontinent exhibit many lexical and syntactic similarities on account of genetic and contact relatedness (Abbi, 2012; Subbārāo, 2012). Genetic relatedness manifests in the two major language groups considered in this work: the Indo-Aryan branch of the Indo-European family and the Dravidian family. Owing to the long history of contact between these language groups, the Indian subcontinent is a *linguistic area* (Emeneau, 1956) exhibiting convergence of many linguistic properties between languages of these groups. Hence, we explore multilingual models spanning all these Indic languages to enable transfer from high resource to low resource languages on account of genetic relatedness (Nguyen and Chiang, 2017) or contact relatedness (Goyal et al., 2020). We trained 2 types of multilingual models for translation involving Indic languages: (i) One to Many for English to Indic language translation (O2M: 11 pairs) (ii) Many to One for Indic language to English translation (M2O: 11 pairs).

**Data Representation.** We made a design choice to represent all the Indic language data in a single script (using the Indic NLP Library). The scripts for these Indic languages are all derived from the ancient Brahmi script. Though each of these scripts have their own Unicode codepoint range, it is possible to get a 1-1 mapping between characters in these different scripts since the Unicode standard takes into account the similarities between these scripts. Hence, we convert all the Indic data to the Devanagari script. This allows better lexical sharing between languages for transfer learning, prevents fragmentation of the subword vocabulary between Indic languages and allows using a smaller subword vocabulary.

The first token of the source sentence is a special token indicating the source language (Tan et al., 2019; Tang et al., 2020). The model can make a decision on the transfer learning between these languages based on both the source language tag and the similarity of representations. When multiple target languages are involved, we follow the standard approach of using a special token in the input sequence to indicate the target language (Johnson et al., 2017b). Other standard pre-processing done on the data are Unicode normalization and tokenization. When the target language is Indic, the output in Devanagari script is converted back to the corresponding Indic script.

**Training Data.** We use all the *Samanantar* parallel data between English and Indic languages and remove overlaps with any test or validation data using a very strict criteria. For the purpose of overlap identification only, we work with lower-cased data with all punctuation characters removed. We remove any translation pair, $(en, t)$, from the training data if (i) the English sentence $en$ appears in the validation/test data of any *En-X* language pair or (ii) the Indic sentence $t$ appears in the validation/test data of the corresponding *En-X* language pair. Note that, since we train a joint model it is important to ensure that no $en$ sentence in the test/validation data appears in any of the *En-X* training sets. For instance, if there is an $en$ sentence in the En-Hi validation/test data then any pair containing this sentence should not be in any of the *En-X* training sets. . We do not use any data sampling while training and leave the exploration of these strategies for future work (Arivazhagan et al., 2019).

**Validation Data.** We used all the validation data from the benchmarks described in Section 5.1.

**Vocabulary.** We learn separate vocabularies for English and Indic languages from English-centric training data using 32K BPE merge operations each using subword-nmt (Sennrich et al., 2016b).

**Network & Training.** We use fairseq (Ott et al., 2019) for training transformer-based models. We use 6 encoder and decoder layers, input embeddings of size 1536 with 16 attention heads and feedforward dimension of 4096. We optimized the cross entropy loss using the Adam optimizer with a label-smoothing of 0.1 and gradient clipping of 1.0. We use mixed precision training with Nvidia Apex[19]. We use an initial learning rate of 5e-4, 4000 warmup steps and the learning rate annealing schedule as proposed in Vaswani et al. (2017). We use a global batch size of 64k tokens. We train each model on 8 V100 GPUs and use early stopping with the patience of 5 epochs.

**Decoding.** We use beam search with a beam size of 5 and length penalty set to 1.

## 5 Experimental Setup

We evaluate the usefulness of *Samanantar* by comparing the performance of a translation system trained using it with existing state of the art models on a wide variety of benchmarks.

### 5.1 Benchmarks

We use the following publicly available benchmarks for evaluating all the models: WAT2020 Indic task (Nakazawa et al., 2020), WAT2021 MultiIndicMT task[20], WMT test sets (Bojar et al., 2014) (Barrault et al., 2019) (Barrault et al., 2020), UFAL Entam (Ramasamy et al., 2012) and the recently released FLORES test set (Goyal et al., 2021). We also create a testset consisting of 1000 validation

[19] https://github.com/NVIDIA/apex

[20] https://lotus.kuee.kyoto-u.ac.jp/WAT/WAT2021/index.html

and 2000 test samples for the en-as pair from PMIndia corpus (Haddow and Kirefu, 2020).

### 5.2 Evaluation Metrics

We use BLEU scores for evaluating the models. To ensure consistency and reproducibility across the models, we provide SacreBLEU signatures in the footnote for Indic-English[21] and English-Indic[22] evaluations. For Indic-English, we use the in-built, default `mteval-v13a` tokenizer. For En-Indic, since SacreBLEU tokenizer does not support Indic languages[23], we first tokenize using the IndicNLP tokenizer before running SacreBLEU. The evaluation script will be made available for reproducibility.

### 5.3 Models

We compare the the following models:

**Commercial MT systems.** We use the translation APIs provided by Google Cloud Platform (v2) (`GOOG`) and Microsoft Azure Cognitive Services (v3) (`MSFT`) to translate all the sentences in the test set of the benchmarks described above.

**Publicly available MT systems.** We consider the following publicly available NMT systems:

*OPUS-MT*[24](`OPUS`): These models were trained using all parallel sources available from OPUS as described in section 2.1. We refer the readers to (Tiedemann and Thottingal, 2020b) for further details about the training data.

*mBART50*[25](`mBART`): This is a multilingual many-to-many model which can translate between any pair of 50 languages. This model is first pre-trained on large amounts of monolingual data from all the 50 languages and then jointly fine-tuned using parallel data between multiple language pairs. We refer the readers to the original paper for details of the monolingual pre-training data and the bilingual fine-tuning data (Tang et al., 2020).

**Models trained on all existing parallel data.** To evaluate the usefulness of the parallel sentences in *Samanantar*, we train a few well studied models using all parallel data available prior to this work.

*Transformer*(`TF`): We train one transformer model each for every en-Indic language pair and one for every Indic-en language pair (22 models in all).

We follow Transformer$_{BASE}$ model described in (Vaswani et al., 2017). We use byte pair encoding (BPE) with a vocabulary size of ≈32K for every language. We use the same learning rate schedule as proposed in (Vaswani et al., 2017). We train each model on 8 V100 GPUs and use early stopping with the patience set to 5 epochs.

*mT5*(`mT5`): We finetune the pre-trained mT5$_{BASE}$ model (Xue et al., 2021) for the translation task using all existing sources of parallel data. We finetune one model for every language pair of interest (18 pairs). We train each model on 1 v3 TPU and use early stopping with a patience of 25K steps.

**Models trained using *Samanantar*** (`IT` [26]). We train the proposed *IndicTrans* model from scratch using the entire *Samanantar* corpus.

For all the models trained/finetuned as a part of this work, we ensured that there is no overlap between the training set and the test/validation sets.

## 6 Results and Discussion

The results of our experiments on Indic-En and En-Indic translation are reported in Table 6 and Table 7. Below, we list down the main observations from our experiments.

**Compilation of existing resources was a fruitful exercise.** We observe that current state-of-the-art models trained on all existing parallel data (curated as a subset of *Samanantar*) perform competitively with other models.

***IndicTrans* trained on *Samanantar* outperforms all publicly available open source models.** From Tables 6 and 7, we observe that *IndicTrans* trained on *Samanantar* outperforms nearly all existing models for all the languages in both the directions. In all cases, except for languages in the WMT and UFAL en-ta benchmark, IndicTrans trained on *Samanantar* improves upon all existing systems. The absolute gain in BLEU score is higher for the Indic-En direction as compared to the En-Indic direction. This is on account of better transfer in many to one settings compared to one-to-many settings (Aharoni et al., 2019) and better language model on the target side. In particular, in Table 7, we observe that *IndicTrans* trained on *Samanantar* clearly outperforms *IndicTrans* trained only on existing resources. Note that the results reported in Table 7 are on the FLORES test set which is a more balanced test set in comparison to the other test sets in Table 6 which are primarily from NEWS sources and have similar distributions as the corresponding training sets. The good performance of our model trained on *Samanantar* on the independently created FLORES test set clearly demonstrates the utility of *Samanantar* in improving the performance of MT models on a wide variety of domains.

[21] BLEU+case.mixed+numrefs.1+smooth.exp+tok.13a+version.1.5.1

[22] BLEU+case.mixed+numrefs.1+smooth.exp+tok.none+version.1.5.1

[23] We plan to submit a pull request in sacrebleu for indic tokenizers

[24] https://huggingface.co/Helsinki-NLP

[25] https://huggingface.co/transformers/model_doc/mbart.html

[26] IT is trained on Samanantar-v0.3 Corpus

| Model | x-en | | | | | | | | | en-x | | | | | | | | |
|---|---|---|---|---|---|---|---|---|---|---|---|---|---|---|---|---|---|---|
| | GOOG | MSFT | CVIT | OPUS | mBART | TF | mT5 | IT | Δ | GOOG | MSFT | CVIT | OPUS | mBART | TF | mT5 | IT | Δ |
| **WAT2021** | | | | | | | | | | | | | | | | | | |
| **bn** | 20.6 | 21.8 | - | 11.4 | 4.7 | 24.2 | 24.8 | **29.6** | 4.8 | 7.3 | 11.4 | 12.2 | - | 0.5 | 13.3 | 13.6 | **15.3** | 1.7 |
| **gu** | 32.9 | 34.5 | - | - | 6.0 | 33.1 | 34.6 | **40.3** | 5.7 | 16.1 | 22.4 | 22.4 | - | 0.7 | 21.9 | 24.8 | **25.6** | 0.8 |
| **hi** | 36.7 | 38.0 | - | 13.3 | 33.1 | 38.8 | 39.2 | **43.9** | 4.7 | 32.8 | 34.3 | 34.3 | 11.4 | 27.7 | 35.9 | 36.0 | **38.6** | 2.6 |
| **kn** | 24.6 | 23.4 | - | - | - | 23.5 | 27.8 | **36.4** | 8.6 | 12.9 | 16.1 | - | - | - | 12.1 | 17.3 | **19.1** | 1.8 |
| **ml** | 27.2 | 27.4 | - | 5.7 | 19.1 | 26.3 | 26.8 | **34.6** | 7.3 | 10.6 | 7.6 | 11.4 | 1.5 | 1.6 | 11.2 | 7.2 | **14.7** | 3.3 |
| **mr** | 26.1 | 27.7 | - | 0.4 | 11.7 | 26.7 | 27.6 | **33.5** | 5.9 | 12.6 | 15.7 | 16.5 | 0.1 | 1.1 | 16.3 | 17.7 | **20.1** | 2.4 |
| **or** | 23.7 | 27.4 | - | - | - | 23.7 | - | **34.4** | 7.0 | 10.4 | 14.6 | 16.3 | - | - | 14.8 | - | **18.9** | 2.6 |
| **pa** | 35.9 | 35.9 | - | 8.6 | - | 36.0 | 37.1 | **43.2** | 6.1 | 22 | 28.1 | - | - | - | 29.8 | 31. | **33.1** | 2.1 |
| **ta** | 23.5 | 24.8 | - | - | 26.8 | 28.4 | 27.8 | **33.2** | 4.8 | 9.0 | 11.8 | 11.6 | - | 11.1 | 12.5 | 13.2 | **13.5** | 0.3 |
| **te** | 25.9 | 25.4 | - | - | 4.3 | 26.8 | 28.5 | **36.2** | 7.7 | 7.6 | 8.5 | 8.0 | - | 0.6 | 12.4 | 7.5 | **14.1** | 1.7 |
| **WAT2020** | | | | | | | | | | | | | | | | | | |
| **bn** | 17.0 | 17.2 | 18.1 | 9.0 | 6.2 | 16.3 | 16.4 | **20.0** | 1.9 | 6.6 | 8.3 | 8.5 | - | 0.9 | 8.7 | 9.3 | **11.4** | 2.1 |
| **gu** | 21.0 | 22.0 | 23.4 | - | 3.0 | 16.6 | 18.9 | **24.1** | 0.7 | 10.8 | 12.8 | 12.4 | - | 0.5 | 9.7 | 11.8 | **15.3** | 2.5 |
| **hi** | 22.6 | 21.3 | 23.0 | 8.6 | 19.0 | 21.7 | 21.5 | **23.6** | 0.6 | 16.1 | 15.6 | 16.0 | 6.7 | 13.4 | 17.4 | 17.3 | **20.0** | 2.6 |
| **ml** | 17.3 | 16.5 | 18.9 | 5.8 | 13.5 | 14.4 | 15.4 | **20.4** | 1.5 | 5.6 | 5.5 | 5.3 | 1.1 | 1.5 | 5.2 | 3.6 | **7.2** | 1.6 |
| **mr** | 18.1 | 18.6 | 19.5 | 0.5 | 9.2 | 15.3 | 16.8 | **20.4** | 0.9 | 8.7 | 10.1 | 9.6 | 0.2 | 1.0 | 9.8 | 10.9 | **12.7** | 1.8 |
| **ta** | 14.6 | 15.4 | 17.1 | - | 16.1 | 15.3 | 14.9 | **18.3** | 1.3 | 4.5 | 5.4 | 4.6 | - | 5.5 | 5.0 | 5.2 | **6.2** | 0.7 |
| **te** | 15.6 | 15.1 | 13.7 | - | 5.1 | 12.1 | 14.2 | **18.5** | 2.9 | 5.5 | 7.0 | 5.6 | - | 1.1 | 5.0 | 5.4 | **7.6** | 0.7 |
| **WMT** | | | | | | | | | | | | | | | | | | |
| **hi** | 31.3 | 30.1 | 24.6 | 13.1 | 25.7 | 25.3 | 26.0 | **29.7** | -1.6 | 24.6 | 24.2 | 20.2 | 7.9 | 18.3 | 23. | 23.8 | **25.5** | 0.9 |
| **gu** | 30.4 | 29.9 | 24.2 | - | 5.6 | 16.8 | 21.9 | **25.1** | -5.4 | 15.2 | 17.5 | 12.6 | - | 0.5 | 9.0 | 12.3 | **17.2** | -0.3 |
| **ta** | 27.5 | 27.4 | 17.1 | - | 20.7 | 16.6 | 17.5 | **24.1** | -3.4 | 9.6 | 10.0 | 4.8 | - | 6.3 | 5.8 | 7.1 | **9.9** | -0.1 |
| **UFAL** | | | | | | | | | | | | | | | | | | |
| **ta** | 25.1 | 25.5 | 19.9 | - | 24.7 | 26.3 | 25.6 | **30.2** | 3.9 | 7.7 | 10.1 | 7.2 | - | 9.2 | 11.3 | **11.9** | 10.9 | -1.0 |
| **PMI** | | | | | | | | | | | | | | | | | | |
| **as** | - | 16.7 | - | - | - | 7.4 | - | **29.9** | 13.2 | - | 10.8 | - | - | - | 3.5 | - | **11.6** | 0.8 |

Table 6: BLEU scores for En-X and X-En translation across different available testsets. Δ represents the difference between IndicTrans and the best results from the other models. We bold the best public model and underline the overall best model.

| Model | x-en | | | | | | | en-x | | | | | | |
|---|---|---|---|---|---|---|---|---|---|---|---|---|---|---|
| | GOOG | MSFT | CVIT | OPUS | mBART | IT† | IT | GOOG | MSFT | CVIT | OPUS | mBART | IT† | IT |
| as | - | 24.9 | - | - | - | 17.1 | **23.3** | - | 13.6 | - | - | - | **7.0** | 6.9 |
| bn | 34.6 | 31.2 | - | 17.9 | 9.4 | 30.1 | **32.2** | 28.1 | 22.9 | 7.9 | - | 1.4 | 18.2 | **20.3** |
| gu | 40.2 | 35.4 | - | - | 4.8 | 30.6 | **34.3** | 25.6 | 27.7 | 14.1 | - | 0.7 | 19.4 | **22.6** |
| hi | 44.2 | 36.9 | - | 18.6 | 32.6 | 34.3 | **37.9** | 38.7 | 31.8 | 25.7 | 13.7 | 22.2 | 32.2 | **34.5** |
| kn | 32.2 | 30.5 | - | - | - | 19.5 | **28.8** | 32.6 | 22.0 | - | - | - | 9.9 | **18.9** |
| ml | 34.6 | 34.1 | - | 9.5 | 24.0 | 26.5 | **31.7** | 27.4 | 21.1 | 6.6 | 4.4 | 3.0 | 10.9 | **16.3** |
| mr | 36.1 | 32.7 | - | 0.6 | 14.8 | 27.1 | **30.8** | 19.8 | 18.3 | 8.5 | 0.1 | 1.2 | 12.7 | **16.1** |
| or | 31.7 | 31.0 | - | - | - | 26.1 | **30.1** | 24.4 | 20.9 | 7.9 | - | - | 11.0 | **13.9** |
| pa | 39.0 | 35.1 | - | 9.9 | - | 30.3 | **35.8** | 27.0 | 28.5 | - | - | - | 21.3 | **26.9** |
| ta | 31.9 | 29.8 | - | - | 22.3 | 24.2 | **28.6** | 28.0 | 20.0 | 7.9 | - | 8.7 | 10.2 | **16.3** |
| te | 38.8 | 37.3 | - | - | 15.5 | 29.0 | **33.5** | 30.6 | 30.5 | 8.2 | - | 4.5 | 17.7 | **22.0** |

Table 7: BLEU scores for En-X and X-En translation for FLORES devtest Benchmark. IT† is IndicTrans trained only on existing data. We bold the best public model and underline the overall best model.

***IndicTrans* trained on *Samanantar* outperforms commercial systems on most datasets.** From Table 6, we observe that *IndicTrans* trained on *Samanantar* outperforms commercial models (GOOG and MSFT) on most benchmarks. On the FLORES dataset our models are still a few points behind the commercial systems. The higher performance of the commercial NMT systems on the FLORES dataset indicates that the in-house training datasets for these systems better capture the domain and data distributions of the FLORES dataset.

**Performance gains are higher for low resource languages.** We observe significant gains for low resource languages such as, or and kn, especially in the Indic-En direction. These languages benefit from other related languages with more resources due to multilingual training.

**Pre-training needs further investigation.** mT5 which is pre-trained on large amounts of monolingual corpora from multiple languages does not always outperform a Transformer$_{BASE}$ model which is just trained on existing parallel data without any pre-training. While this does not invalidate the value of pre-training, it does suggest that pre-training needs to be optimized for the specific languages. As future work, we would like to explore pre-training using the monolingual corpora on Indic languages available from IndicCorp. Further, we would like to pre-train a single script mT5- or mBART-like model for Indic languages and then fine-tune on MT using *Samanantar*.

## 7 Conclusion

We present *Samanantar*, the largest publicly available collection of parallel corpora for Indic languages. In particular, we mine 37.4M parallel sentences by leveraging web crawled monolingual corpora as well as recent advances in multilingual representation learning, approximate nearest neighbor search, and optical character recognition. We also mine 83.4M parallel sentences between 55 Indic language pairs from this English-centric corpus. We collect human judgments for 9,566 sentence pairs from *Samanantar* and show that the newly mined pairs are of high quality. Our multilingual single-script model, *IndicTrans*, trained on *Samanantar* outperforms existing models on a wide variety of benchmarks, demonstrating that our parallel corpus mining approaches can contribute to high-quality MT models for Indic languages.

To further improve the parallel corpora and translation quality for Indian languages, the following areas need further exploration: (a) improving LaBSE representations for low-resource languages and longer sentences, especially benefiting from human judgments, (b) optimising training schedules and objectives such that they utilize data quality information and linguistic similarity, (c) pre-training multilingual models.

We hope that the three main contributions of this work, *viz.*, *Samanantar*, *IndicTrans* and a manually annotated dataset for cross-lingual similarity will contribute to further research on NMT and multilingual NLP for Indic languages.

## Acknowledgements

We would like to thank the TACL editors and reviewers, who have helped us shape this paper. We would like to thank EkStep Foundation for their generous grant which went into hiring human resources as well as cloud resources needed for this work. We would like to thank the Robert Bosch Center for Data Science and Artificial Intelligence for supporting Sumanth and Gowtham through their Post Baccalaureate Fellowship Program. We would like to thank Google for their generous grant through their TPU Research Cloud Program. We would also like to thank the following members from Tarento Technlogies for providing logistical and technical support: Sivaprakash Ramasamy, Amritha Devadiga, Karthickeyan Chandrasekar, Naresh Kumar, Dhiraj D, Vishal Mahuli, Dhanvi Desai, Jagadeesh Lachannagari, Dhiraj Suthar, Promodh Pinto, Sajish Sasidharan, Roshan Prakash Shah, Abhilash Seth.